\RequirePackage[svgnames]{xcolor}

\documentclass[11pt,letterpaper]{mystyle}
\usepackage{lmodern}
\usepackage{tikz}
\usetikzlibrary{
  shapes,               
  arrows.meta,          
  positioning,          
  decorations.pathreplacing, 
  fit                   
}

\usepackage{graphicx}
\usepackage{wrapfig}
\usepackage[all]{hypcap}
\usepackage[svgnames]{xcolor}
\usepackage[comma,authoryear,compress]{natbib}
\usepackage{hyperref}[citecolor=lightblue]
\usepackage{makecell}
\hypersetup{
    colorlinks = true,
    citecolor = {YaleBlue},
}
\usepackage{graphicx}
\usepackage{subcaption}

\usepackage{algorithm}
\usepackage{algorithmicx}
\usepackage{algpseudocode}
\usepackage{caption} 
\usepackage{microtype}
\usepackage{graphicx}
\expandafter\def\csname ver@subfig.sty\endcsname{}
\usepackage{booktabs} %
\usepackage{float}
\usepackage{bigstrut}
\usepackage{booktabs}
\usepackage{amsmath}
\usepackage{amssymb}
\usepackage{mathtools}
\usepackage{amsthm}
\usepackage{mathrsfs}
\usepackage{nicefrac}
\usepackage{dsfont}
\usepackage{enumitem}

\usepackage{cleveref}
\usepackage{bxcoloremoji}
\usepackage{epigraph}
\usepackage{float}

\usepackage{arydshln}
\usepackage[utf8]{inputenc} %
\usepackage[T1]{fontenc}    %
\usepackage{hyperref}       %
\usepackage{url}            %
\usepackage{booktabs}       %
\usepackage{amsfonts}       %
\usepackage{nicefrac}       %
\usepackage{microtype}      %
\usepackage{graphicx}
\usepackage{enumitem}

\usepackage{amssymb}
\usepackage{fdsymbol}
\usepackage{wrapfig}
\usepackage{lipsum}
\usepackage{enumitem}
\usepackage{stackengine}
\usepackage[font=small,labelfont=bf]{caption}

\usepackage{rotating}
\usepackage{makecell}
\usepackage{multirow}
\usepackage{tabularx}
\newcolumntype{Y}{>{\centering\arraybackslash}X}

\usepackage{bera}
\usepackage{listings}
\usepackage{xcolor}

\definecolor{eclipseStrings}{RGB}{42,0,255}
\definecolor{eclipseKeywords}{RGB}{127,0,85}
\colorlet{numb}{magenta!60!black}
\definecolor{bggray}{rgb}{0.95, 0.95, 0.95}
\definecolor{linenobg}{rgb}{1,1,1}  

\lstdefinelanguage{json}{
    basicstyle=\footnotesize\ttfamily,
    commentstyle=\color{eclipseStrings},
    stringstyle=\color{eclipseKeywords},
    numbers=left,
    numberstyle=\scriptsize\color{black}\setlength{\fboxsep}{0pt}\colorbox{linenobg},
    stepnumber=1,
    numbersep=10pt,
    xleftmargin=2em,
    framexleftmargin=2em,
    showstringspaces=false,
    breaklines=true,
    frame=single,
    backgroundcolor=\color{bggray},
    string=[s]{"}{"},
    comment=[l]{:},
    morecomment=[l]{,},
}

\definecolor{dockerbg}{rgb}{0.97, 0.97, 0.97}
\definecolor{scriptbg}{rgb}{0.90, 0.95, 1.00}  

\PassOptionsToPackage{most}{tcolorbox}
\usepackage[most]{tcolorbox}

\usepackage[most]{tcolorbox}
\tcbuselibrary{listings,breakable,minted}
\usepackage{alltt}
\definecolor{blue1}{HTML}{196ab1}
\definecolor{blue2}{HTML}{4886c1}
\definecolor{blue3}{HTML}{5e9bd6}
\definecolor{blue4}{HTML}{77b1e2}
\definecolor{blue5}{HTML}{bdd930}
\definecolor{blue6}{HTML}{dfebf6}

\definecolor{red1}{HTML}{de512c}
\definecolor{red2}{HTML}{f2642d}
\definecolor{red3}{HTML}{f68f58}
\definecolor{red4}{HTML}{febf92}
\definecolor{red5}{HTML}{f8e9c8}
\tcbset{
  images/.style={
    colback=dockerbg,                   
    colframe=gray,           
    boxrule=1pt,                     
    arc=0mm,                         
    colbacktitle=gray,       
    coltitle=white,                  
    fonttitle=\bfseries\sffamily,    
    title={TITLE},   
    toptitle=2mm,                   
    bottomtitle=2mm,                
    leftrule=2pt, rightrule=2pt,     
    top=1mm, bottom=1mm, left=1mm, right=1mm, 
  },
  shell/.style={
    colback=blue!5,                   
    colframe=blue,           
    boxrule=1pt,                     
    arc=2mm,                         
    colbacktitle=blue,       
    coltitle=white,                  
    fonttitle=\bfseries\sffamily,    
    title={TITLE},   
    toptitle=2mm,                   
    bottomtitle=2mm,                
    leftrule=2pt, rightrule=2pt,     
    top=1mm, bottom=1mm, left=3mm, right=3mm, 
  }
}

\setlist[itemize]{topsep=0.3em, partopsep=0pt, parsep=0pt, itemsep=0.4em}

\newcommand{\x}{\mathbf{x}}

\definecolor{blanchedalmond}{rgb}{1.0, 0.92, 0.8}
\definecolor{carmine}{rgb}{0.59, 0.0, 0.09}
\definecolor{lightblue}{rgb}{0.22,0.45,0.70}%

\renewcommand{\mathbf}{\boldsymbol}

\makeatletter
\def\Ddots{\mathinner{\mkern1mu\raise\p@
\vbox{\kern7\p@\hbox{.}}\mkern2mu
\raise4\p@\hbox{.}\mkern2mu\raise7\p@\hbox{.}\mkern1mu}}
\makeatother

\definecolor{amaranth}{rgb}{0.9, 0.17, 0.31}
\definecolor{antiquebrass}{rgb}{0.8, 0.58, 0.46}
\definecolor{antiquefuchsia}{rgb}{0.57, 0.36, 0.51}
\definecolor{chromeyellow}{rgb}{0.31, 0.47, 0.26}

\newtcolorbox{AIbox}[2][]{aibox,title=#2,#1}
\definecolor{lightblue}{rgb}{0.22,0.45,0.70}%
\definecolor{Gray}{gray}{0.95}
\definecolor{Cornsilk}{rgb}{1.0, 0.97, 0.86}

\usepackage{amsmath}
\usepackage{ragged2e}

\usepackage[all]{hypcap}
\usepackage{siunitx}
\title{Riemann-1.0: An Embodied World Action Model for Physical AI}
\paperauthors{\small
Haofeng Sun$^*$, Jiangbo Pei$^*$, Fei Kang$^*$, Zexiang Liu$^*$, Yaokun Li$^*$, Boyi Jiang$^*$, Hua Xue$^*$, Cindy Zhou{$^\ddagger$},\\ Wei Li, Yichen Wei, Mengyin An, Fanliang Zhao, Biao Jiang, Zile Wang, Yang Liu{$^\dagger$}, Yangguang Li{$^\dagger$}\\[0.3em]

\textbf{Riemann Dynamics}\\
research@riemanndynamics.ai
}

\author{}
\date{}

\runningtitle{Riemann-1.0}

\begin{document}

\begin{abstract}

We introduce \textbf{Riemann-1.0}, a fully causal autoregressive World Action Model for embodied intelligence. Riemann-1.0 jointly models multi-view visual observations, robot states, and embodiment-specific actions within a unified causal autoregressive sequence, representing robot actions and world evolution as causal state transitions. Unlike existing WAMs based on joint generation, video-first prediction, or decoupled modeling paradigms, Riemann-1.0 unifies online robot policy execution and action-conditioned world simulation within a single model, enabling it to function as both an executable robot policy and a multi-embodiment visual world simulator. To scale embodied experience across heterogeneous data sources, we further develop a progressive embodied pretraining framework that unifies learning from egocentric human videos, handheld-gripper demonstrations, and heterogeneous robot trajectories under a shared World Action Modeling objective. Built upon \num{200}K+ hours of interaction data, Riemann-1.0 progressively transfers large-scale embodied experience into executable robot manipulation capabilities. Riemann-1.0 achieves state-of-the-art performance across both simulation benchmarks and real-world manipulation tasks. It achieves success rates of 94.3\% on RoboTwin2.0, 99.0\% on LIBERO, and 62.6\% on the long-horizon compositional benchmark RoboCasa-365, outperforming the previous best method by 8.4\%. On long-horizon real-world manipulation tasks, Riemann-1.0 achieves a Success Rate (SR) of 85.0\% and a Progress Success Rate (PSR) of 94.4\%, exceeding the strongest open-source baseline by 15\% in SR. These results demonstrate that unified World Action Modeling together with progressive embodied pretraining effectively transforms large-scale embodied experience into generalizable robot manipulation capabilities.


\textbf{Website}: \href{}{\textcolor{blue}{https://riemann-dynamics.github.io/Riemann-1.0-Website}}



\end{abstract}

\renewcommand{\thefootnote}{\fnsymbol{footnote}}
\footnotetext[1]{Equal Contribution.}
\footnotetext[3]{Research Intern at Riemann Dynamics.}
\footnotetext[2]{Project Lead and Corresponding Author.}
\renewcommand{\thefootnote}{\arabic{footnote}}

\maketitle

\vspace{-3em}
\section{Introduction}

\begin{figure}
    \centering
    \includegraphics[width=1\textwidth]{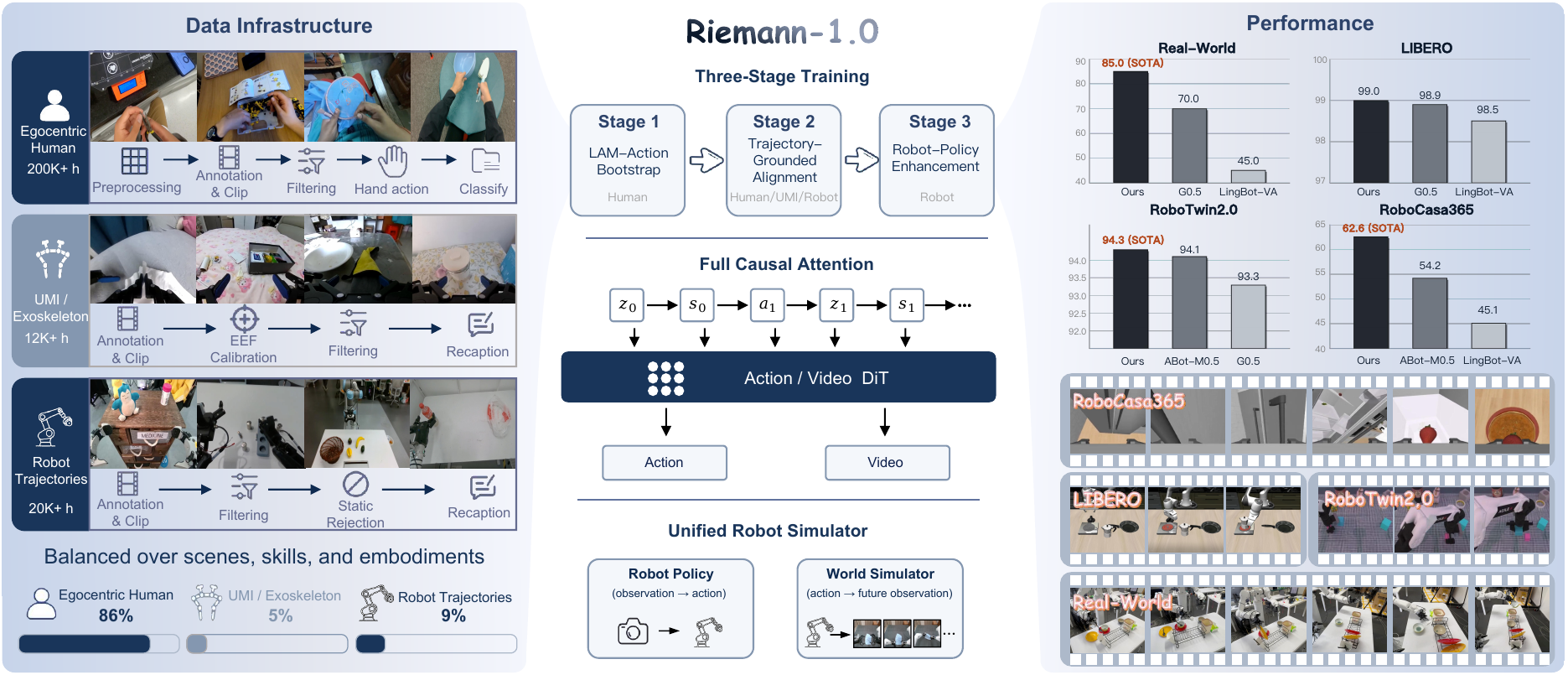}
    \caption{\textbf{Overview of Riemann-1.0}. Riemann-1.0 integrates a unified embodied data infrastructure, Progressive Embodied Pretraining, and a fully causal World Action Model into a unified framework for embodied intelligence. The resulting model simultaneously serves as an executable robot policy and an action-conditioned visual world simulator, achieving state-of-the-art performance across both simulation benchmarks and long-horizon real-world manipulation tasks.}
    \label{fig:teaser}
\end{figure}

Physical AI is driving artificial intelligence beyond understanding the digital world toward perceiving, reasoning, and acting in the physical world. Building intelligent agents for real-world environments requires Embodied Brain Models that unify environmental perception, world dynamics modeling, executable action generation, and long-horizon reasoning. Similar to how Large Language Models continuously improve by scaling textual knowledge, Embodied Brain Models should continuously evolve by scaling embodied experience. In recent years, World Action Models (WAMs) have emerged as an promising paradigm for jointly modeling robot actions and world dynamics~\citep{ye2026dreamzero,ye2026gigaworld,li2026lingbotva,yuan2026fastwam} , providing a unified framework for robot policy learning, predictive world modeling, and long-horizon planning. Together, these advances establish WAMs as a strong foundation for next-generation embodied intelligence.

However, embodied experience is inherently heterogeneous, multimodal, and embodiment-dependent. Substantial differences in observation modalities, action spaces, and supervision signals across data sources make unified large-scale learning fundamentally challenging. Egocentric human videos capture diverse real-world interactions, rich manipulation skills, and long-horizon task structures. Handheld-gripper demonstrations (e.g., UMI) naturally bridge human interactions with robot action spaces through robot-compatible manipulation, while robot trajectories provide precise, embodiment-specific, and executable control supervision. Rather than being redundant, these complementary data sources provide distinct forms of supervision: egocentric human videos contribute scalable interaction knowledge, handheld-gripper demonstrations bridge cross-embodiment action alignment, and robot trajectories ground embodiment-specific executable control. Motivated by this complementary supervision, we construct one of the largest embodied experience corpora to date, comprising \num{200}K+ hours of data spanning thousands of diverse interaction skills  across egocentric human videos, handheld-gripper demonstrations, and heterogeneous robot trajectories.

Despite the rapid progress of World Action Models, existing approaches remain limited in fully leveraging embodied experience at scale. First, existing pretraining strategies are tightly coupled with robot trajectories and therefore cannot effectively leverage large-scale weakly supervised human data, limiting knowledge transfer across heterogeneous embodiments. Second, existing World Action Model formulations—including joint generation~\citep{ye2026dreamzero}, video-first prediction~\citep{li2026lingbotva}, and decoupled action-video modeling~\citep{yuan2026fastwam}—capture different aspects of robot interaction but fail to jointly model observations, robot states, actions, and world evolution within a unified causal process. Consequently, they cannot simultaneously support efficient robot policy learning and causally consistent action-conditioned world simulation within a unified model.

To address these challenges, we present Riemann-1.0, a Fully Causal Autoregressive World Action Model for embodied intelligence, as shown in Figure~\ref{fig:teaser}. Riemann-1.0 formulates robot interaction as a unified causal autoregressive sequence over multi-view visual observations, robot states, and embodiment-specific actions, where robot actions and world evolution are represented as causal state transitions. This fully causal formulation naturally aligns with the interaction process of real-world robots, enabling a single model to simultaneously function as both an executable robot policy and a multi-embodiment action-conditioned visual world simulator.

To effectively leverage large-scale embodied experience, we further propose Progressive Embodied Pretraining, a three-stage curriculum that progressively transfers embodied knowledge from weak supervision to executable robot control. The proposed curriculum follows a natural progression from weakly supervised embodied knowledge learning to executable action alignment, and finally embodiment-specific policy enhancement, enabling continuous capability transfer from large-scale embodied experience to executable robot manipulation. Specifically, a frozen Latent Action Model (LAM) first converts large-scale unlabeled egocentric videos into latent action representations, providing scalable weak action supervision for pretraining. Subsequently, 3D hand-annotated egocentric videos, handheld-gripper demonstrations, and heterogeneous robot trajectories progressively align latent actions with executable robot actions across heterogeneous embodiments. Finally, high-quality robot trajectories specialize the model for embodiment-specific action generation and downstream robot policy learning.

Extensive experiments demonstrate that Riemann-1.0 achieves state-of-the-art performance across both simulation benchmarks and long-horizon real-world manipulation tasks. Riemann-1.0 achieves success rates of 94.3\% on RoboTwin2.0~\citep{chen2025robotwin}, 99.0\% on LIBERO~\citep{liu2023libero}, and 62.6\% on the long-horizon compositional benchmark RoboCasa-365~\citep{nasiriany2026robocasa365}, outperforming the best method by 8.4 percentage points. On long-horizon real-world manipulation tasks, it achieves 85.0\% Success Rate (SR) and 94.4\% Progress Success Rate (PSR), exceeding the strongest open-source baseline by 15 percentage points in SR. These results demonstrate that combining Fully Causal World Action Modeling with Progressive Embodied Pretraining effectively translates large-scale embodied experience into generalizable long-horizon robot manipulation capabilities.

Our contributions are summarized as follows:
\begin{itemize}
    \item \textbf{Fully Causal Autoregressive World Action Model.}
    We propose a fully causal autoregressive World Action Model that jointly models multi-view visual observations, robot states, and embodiment-specific actions within a unified causal sequence. The proposed formulation unifies executable robot policy learning and action-conditioned visual world simulation within a single model.

    \item \textbf{Progressive Embodied Pretraining.}
    We develop a three-stage Progressive Embodied Pretraining framework that progressively transfers embodied knowledge from weakly supervised human interaction data to executable robot control. Built on more than \num{200}K+ hours of embodied experience covering thousands of interaction skills, the proposed framework effectively unifies heterogeneous embodied experience under a shared World Action Modeling objective.

    \item \textbf{Comprehensive Evaluation.}
    We demonstrate the effectiveness of the proposed framework through extensive evaluations on both simulation benchmarks and long-horizon real-world manipulation tasks. Riemann-1.0 achieves state-of-the-art performance across multiple benchmarks while significantly improving long-horizon real-world robot manipulation.

\end{itemize}

\section{Data Infrastructure}

Scaling World Action Models fundamentally depends on scaling embodied experience rather than robot trajectories alone. However, embodied experience is inherently heterogeneous across egocentric human videos, handheld-gripper demonstrations, and robot trajectories. These sources differ substantially in observation modalities, action representations, temporal granularity, and supervision fidelity, making unified large-scale pretraining fundamentally challenging.

To address this challenge, we build a unified embodied data infrastructure that automatically transforms heterogeneous embodied experience into a common action-level video--state--action representation. The proposed infrastructure consists of three tightly coupled components: multi-source embodied experience collection, a unified embodied data engine, and progressive supervision with multi-source data balancing. Together, they produce a scalable pretraining corpus containing more than \num{200}K+ hours of embodied experience covering thousands of interaction skills, providing the data foundation for World Action Modeling.

\begin{figure}
    \centering
    \includegraphics[width=1\textwidth]{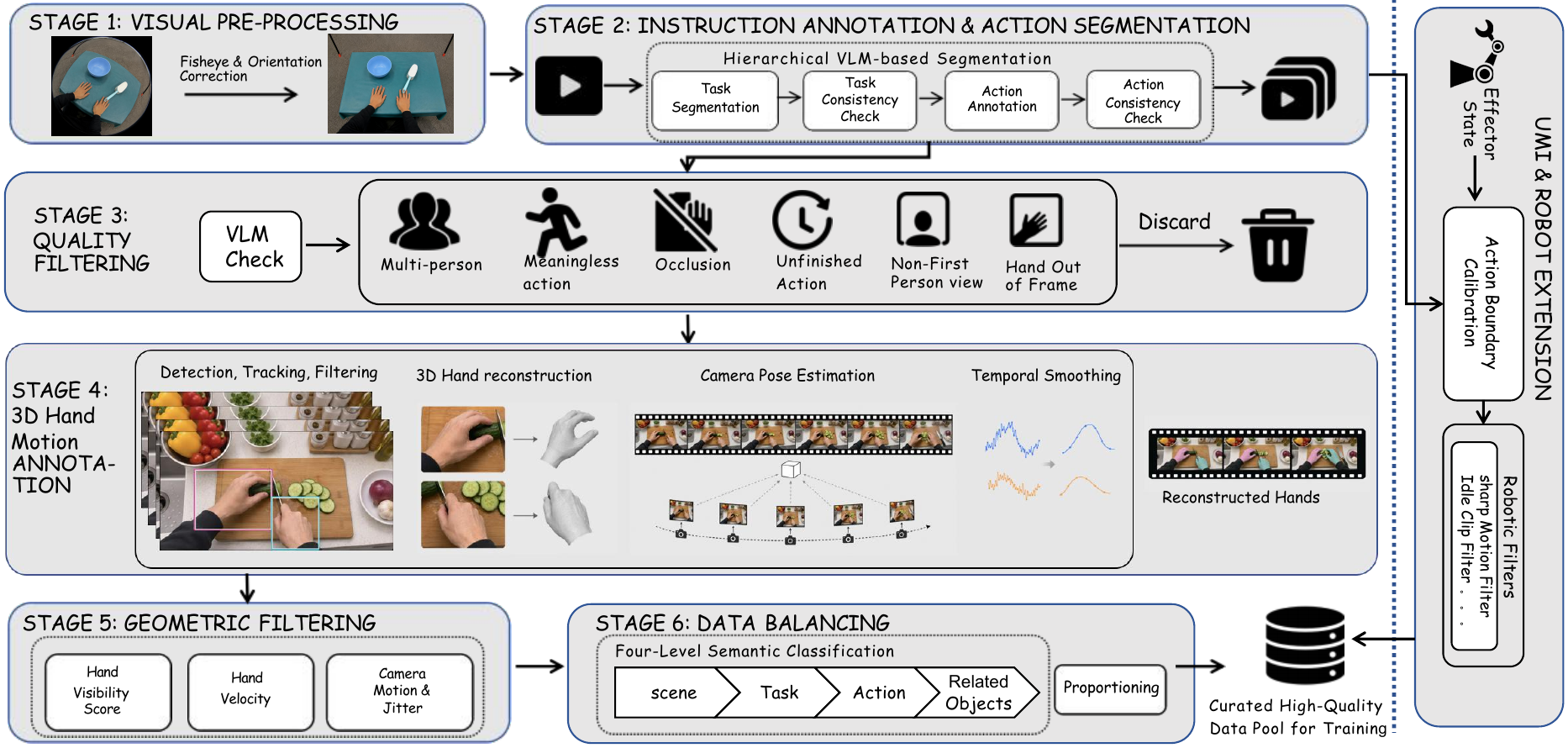}
    \caption{\textbf{Unified embodied data processing pipeline.} The proposed data engine transforms heterogeneous embodied experience into standardized action-level trajectories through six processing stages, including visual preprocessing, semantic annotation, quality filtering, 3D hand reconstruction, geometric filtering, and semantic-aware data balancing. Handheld-gripper demonstrations and robot trajectories are additionally refined through action calibration and robot-specific filtering. Together, these processes produce a unified embodied corpus that supports Progressive Embodied Pretraining and large-scale World Action Modeling.}
    \label{fig:data_process}
\end{figure}

\subsection{Multi-Source Embodied Experience}

We construct a large-scale embodied experience corpus containing more than \num{230}K+ hours of embodied experience, including \num{200}K+ hours of egocentric human videos, \num{12}K+ hours of handheld-gripper and wearable demonstrations, and \num{20}K+ hours of heterogeneous robot trajectories. The corpus spans diverse household, office, and industrial environments, covering thousands of interaction skills and a wide range of robot embodiments.

The three data sources provide complementary supervision. Egocentric human videos offer the broadest diversity of real-world interactions, object-centric manipulation skills, and long-horizon task compositions. Handheld-gripper and wearable demonstrations provide structured end-effector trajectories that naturally bridge human interactions and robot-compatible control. Robot trajectories provide precise embodiment-specific states and directly executable actions for downstream policy learning.

Rather than being redundant, these sources complement each other: human videos provide scalable interaction knowledge, handheld-gripper demonstrations reduce the embodiment gap between humans and robots, and robot trajectories provide executable control supervision. Together, they establish a natural supervision hierarchy for Progressive Embodied Pretraining.

\subsection{Unified Embodied Data Engine}

As illustrated in Figure~\ref{fig:data_process}, although the three data sources provide complementary supervision, their native formats are fundamentally different. Egocentric human videos provide visual observations with weak language supervision, handheld-gripper demonstrations provide structured end-effector trajectories and gripper states, while robot datasets differ substantially in embodiment-specific states, action spaces, coordinate systems, and control frequencies. Directly combining these heterogeneous sources would result in inconsistent supervision and unstable optimization.

To unify heterogeneous embodied experience, we build a unified embodied data engine that converts all data into a common action-level trajectory representation consisting of language instructions, embodiment identities, visual observations, states, actions, and semantic metadata. For each embodiment, state and action streams are temporally aligned, normalized, and organized under embodiment-specific canonical definitions, while validity masks handle variable-dimensional actions without forcing heterogeneous embodiments into a shared physical action space. This unified representation serves as a standardized interface between heterogeneous embodied data and the World Action Modeling objective.

For egocentric human videos, we first perform visual normalization and then apply a hierarchical VLM-based annotation pipeline that progressively decomposes long recordings into task-level and action-level clips with structured semantic annotations, including tasks, actions, scenes, manipulated objects, and temporal boundaries. Verification of consistency at task-level and action-level further improves annotation reliability. To bridge visual interactions and executable actions, we further reconstruct continuous 3D hand trajectories from video-only egocentric human data. Hand detection and tracking first establish temporally consistent hand trajectories, followed by MANO-based~\citep{romero2017embodied} 3D hand reconstruction to recover hand keypoints, wrist poses, and MANO parameters. In parallel, VGGT-$\Omega$~\citep{wang2026vggtomega} estimates camera intrinsics and frame-level camera poses, enabling hand trajectories to be organized in a consistent coordinate system under egocentric camera motion. A lightweight temporal refinement further suppresses jitter and short-term drift, producing temporally continuous hand-action trajectories for large-scale pretraining. Finally, semantic consistency, interaction completeness, hand motion, visibility, and camera motion are jointly evaluated to retain high-quality interaction clips.

Unlike egocentric human videos, handheld-gripper demonstrations and robot trajectories already provide structured control supervision. Their processing therefore mainly focuses on temporal alignment, action normalization, and trajectory quality filtering. Gripper-state transitions are used to refine action boundaries, while camera-shake filtering, stationary end-effector rejection, and abnormal control filtering further improve data quality. After processing, all data sources are converted into temporally aligned action-level video--state--action trajectories under a unified semantic taxonomy, enabling scalable World Action Modeling across heterogeneous embodiments.

\subsection{Progressive Supervision and Data Balancing}

The processed corpus forms a continuum of supervision with progressively increasing action fidelity rather than a simple division between labeled and unlabeled data. Large-scale egocentric human videos first provide broad interaction dynamics and weak action supervision through pseudo actions generated by a frozen Latent Action Model. Human videos with reconstructed 3D hand trajectories, handheld-gripper demonstrations, and heterogeneous robot trajectories subsequently provide continuous real-action supervision for visual-action alignment. Finally, high-quality robot trajectories specialize the model for embodiment-specific action prediction and closed-loop policy execution. This supervision hierarchy naturally matches the three-stage training curriculum of Riemann-1.0, enabling the model to progressively transfer from open-world interaction knowledge to executable robot control.

The raw corpus exhibits severe long-tail distributions across data sources, scenes, tasks, skills, objects, and robot embodiments. Rather than balancing raw data volume, we construct a unified semantic taxonomy and perform semantic-aware sampling across multiple semantic dimensions. This strategy preserves the scale advantage of large human datasets while ensuring sufficient exposure to long-tail manipulation skills and low-resource robot embodiments.

Overall, our data infrastructure transforms heterogeneous embodied experience into a scalable supervision source for World Action Modeling, enabling World Action Models to continuously improve by scaling embodied experience rather than robot trajectories alone.

\section{Riemann-1.0 Design}

\subsection{Preliminary}

WAMs capture robot-environment dynamics by jointly modeling robot actions and their visual consequences.
Given a task condition \(c\), a visual latent \(z_t\), a robot state \(s_t\), and an action \(a_t\), a WAM models the coupled evolution of control and observation over time.

\begin{figure}[t]
    \centering
    \includegraphics[width=\linewidth]{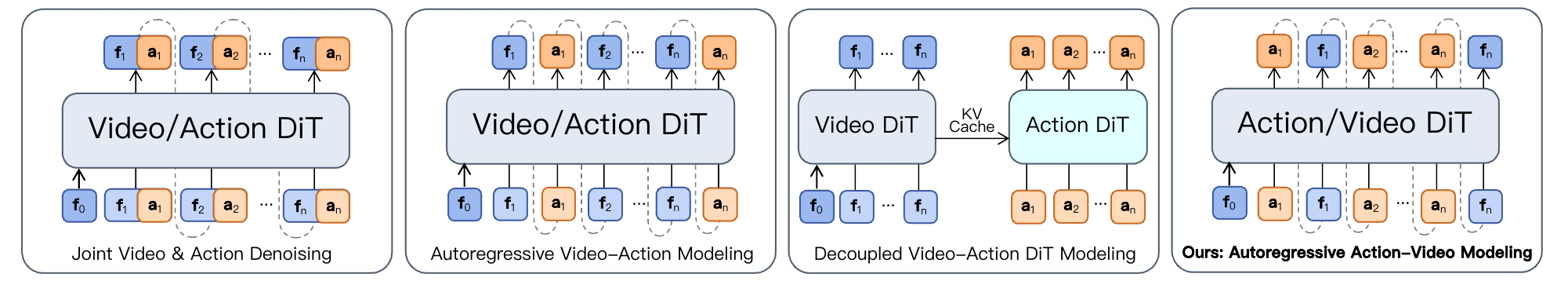}
    \caption{Comparison of joint, video-first, video-action separated DiT, and action-first WAM paradigms.}
    \label{fig:wam_paradigms}
\end{figure}

\paragraph{Flow matching.}
Flow matching learns a continuous vector field that transports noise samples toward data samples. For a data variable \(x\), Gaussian noise \(\epsilon\sim\mathcal{N}(0,I)\), and noise level \(\sigma\in[0,1]\), we define the interpolated sample \(x_\sigma=(1-\sigma)x+\sigma\epsilon\). The target velocity along this path is \(v^\star=\epsilon-x\). A conditional model \(v_\theta\) is trained with \(\mathcal{L}_{\mathrm{fm}}=\mathbb{E}_{x,\epsilon,\sigma}[\|v_\theta(x_\sigma,\sigma,c)-(\epsilon-x)\|_2^2]\).
This formulation naturally fits World Action Modeling, enabling visual latents and continuous robot actions to be optimized within a unified velocity-regression framework.

\paragraph{Comparison of WAM paradigms.}

Existing WAM paradigms primarily differ in how they model the interaction between robot actions and future visual observations, leading to different trade-offs among policy learning, visual simulation, and inference efficiency (Figure~\ref{fig:wam_paradigms}). DreamZero-style methods~\citep{ye2026dreamzero} jointly denoise visual latents and actions as \(p(z_{1:T},a_{1:T}\mid z_0,s_0,c)\). This tightly couples action and visual generation, but requires jointly modeling modalities that have different dimensionalities, temporal resolutions, and optimization characteristics. Video-first methods, such as LingBot-VA~\citep{li2026lingbotva}, first generate future observations and then infer actions from the predicted visual trajectory, i.e., \(p(z_{1:T}\mid z_0,c)\,p(a_{1:T}\mid z_{\leq T},s_{\leq T},c)\), thereby introducing additional inference latency. FastWAM-style methods~\citep{yuan2026fastwam} instead decouple video and action generation by using two separate DiTs, improving modularity while still conditioning action prediction on predicted visual features rather than causal interaction histories. In contrast, Riemann-1.0 adopts a fully causal autoregressive action-video formulation, where robot actions are either predicted online or externally specified before future visual latents are generated. This fully causal formulation enables a single model to function both as an executable robot policy and as an action-conditioned visual world simulator.

\subsection{Riemann-1.0: Fully Causal Action-Video World Model}
\begin{figure}[t]
    \centering
    \includegraphics[width=\linewidth]{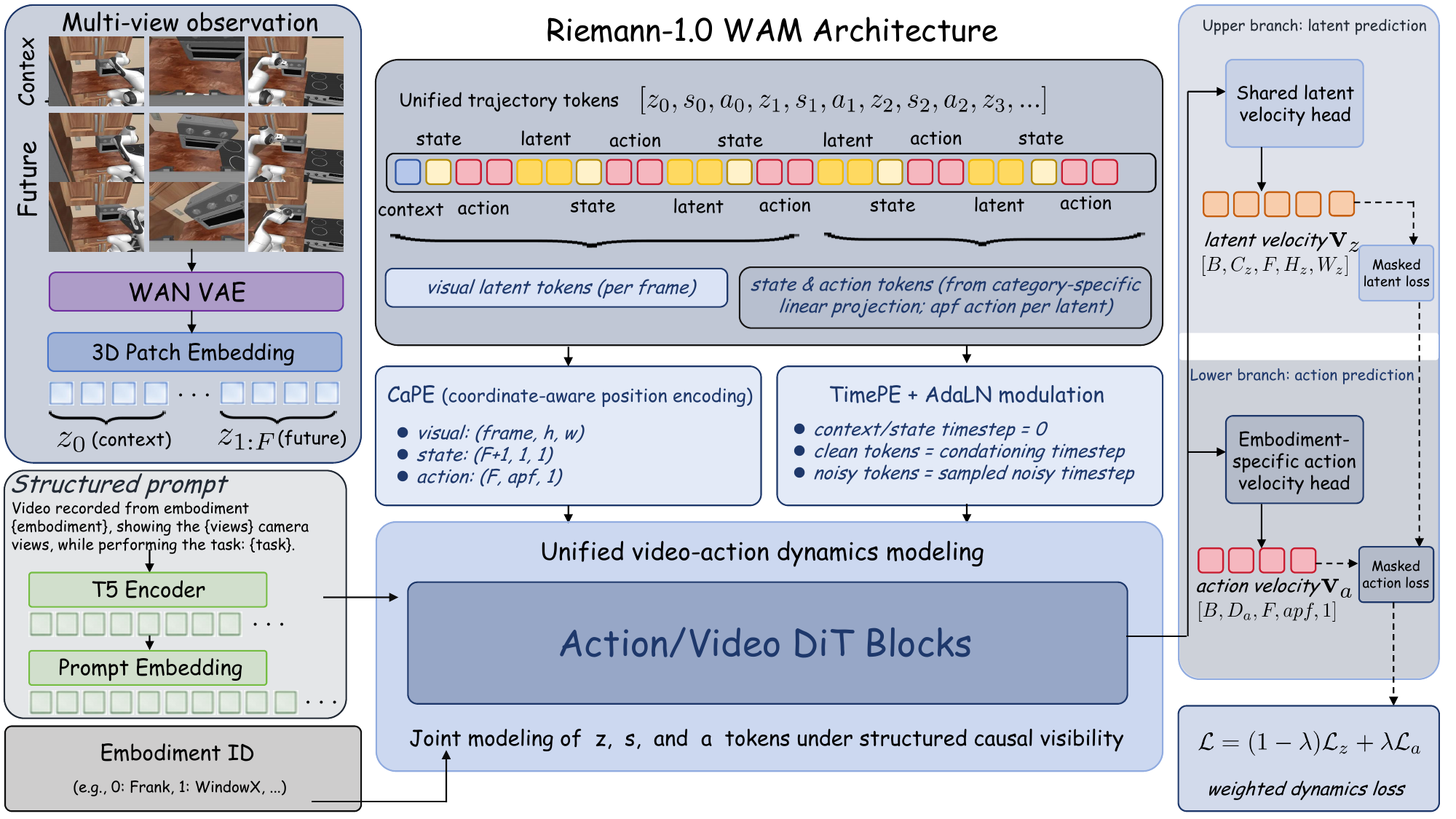}
    \caption{Causal action-video architecture of Riemann-1.0, jointly modeling action prediction and visual latent generation conditioned on task instructions, robot states, and interaction history.}
    \label{fig:rima_wam}
\end{figure}
Riemann-1.0 adopts a fully causal autoregressive factorization over actions and visual latents. As shown in Figure~\ref{fig:rima_wam}, at each autoregressive step, the model predicts the current action from the preceding visual observations, robot states, and action histories. The action is then appended to the causal context and used as a condition for predicting the corresponding visual latent. This ordering follows the causal interaction process of real robots, where actions are executed before their visual consequences are observed. During policy deployment, the predicted visual latent is replaced by the real observation returned by the environment after action execution. During visual simulation, the predicted visual latent is recursively fed back into the model to generate future action-conditioned observations:
\begin{equation}
p(a_{1:T},z_{1:T}\mid z_0,s_0,c)
=
\prod_{t=1}^{T}
p(a_t\mid z_{<t},s_{<t},a_{<t},c)\,
p(z_t\mid z_{<t},s_{<t},a_{\leq t},c).
\label{eq:unified_ar_wam}
\end{equation}
Robot states \(s_t\) are not generated by the model. Instead, they are directly observed from the environment or replayed from recorded trajectories and injected as conditioning signals for the next autoregressive step.
This factorization preserves the causal interaction process: historical observations, robot states, and previous actions determine the next robot action, which in turn conditions the next visual transition. The resulting observation together with the updated robot state then becomes the context for the following autoregressive step.
Compared with GIGA-World-Policy~\citep{ye2026gigaworld}, which inherits its visual dynamics prior from large-scale video generation pretraining and subsequently adapts it for robot policy learning, Riemann-1.0 adopts the same fully causal action-video formulation throughout progressive embodied pretraining, post-training, and deployment. Under a unified World Action Modeling objective, Riemann-1.0 jointly learns latent or executable actions together with visual dynamics from the beginning of pretraining, allowing its dynamics prior to be directly shaped by embodied action supervision rather than transferred from a general-purpose video generation objective. Moreover, each predicted or observed visual consequence is recursively incorporated into the causal interaction history for subsequent action prediction, enabling long-horizon closed-loop interaction. This unified formulation minimizes the modeling mismatch between learned supervision and online execution, allowing the learned action-world dynamics to transfer naturally from large-scale embodied pretraining to downstream robot policy learning.

Eq.~\ref{eq:unified_ar_wam} defines the autoregressive factorization of the joint action-video rollout conditioned on the task prompt \(c\), the initial visual latent \(z_0\), and the initial state \(s_0\). Specifically,
Riemann-1.0 predicts actions from the preceding visual, state, and action histories, and generates each future visual latent from the same causal history together with the current action.
Therefore, an action chunk is not only a policy output, but also a conditioning input to the visual dynamics model, which is the key interface that lets Riemann-1.0 act as a robot world simulator.

\paragraph{Inputs, prompt conditioning, and normalization.}
The language condition specifies the task instruction together with the robot embodiment and camera-view configuration:
\[
\begin{aligned}
&\text{\small\texttt{Video recorded from embodiment \{embodiment\_type\},}}\\
&\text{\small\texttt{showing the \{num\_views\} camera views, while performing the task: \{task\}.}}
\end{aligned}
\]
The language condition is encoded by a T5 encoder~\citep{colin2020exploring} and injected into the transformer through cross-attention, providing semantic task and embodiment context. A numerical embodiment ID is further used to select embodiment-specific action/state projections and prediction heads, enabling heterogeneous robots to share a unified backbone while preserving embodiment-specific control parameterizations.
Visual observations are packed into a unified embodiment-specific multi-view canvas, resized to the model resolution, encoded into latent representations using the Wan VAE~\citep{wan2025}, and converted into transformer tokens through a 3D patch embedding. The first latent \(z_0\) serves as the initial observation context, while future visual latents \(z_{1:T}\) are generated autoregressively.
Actions and robot states are transformed into embodiment-specific canonical representations and normalized using per-embodiment statistics. This normalization removes differences in coordinate systems, physical units, and action dimensionalities while preserving embodiment-specific control semantics, allowing heterogeneous embodiments to share a unified World Action Modeling interface.

\begin{wrapfigure}{r}{0.5\textwidth}
    \centering
    \vspace{-12pt}
    \includegraphics[width=0.5\textwidth]{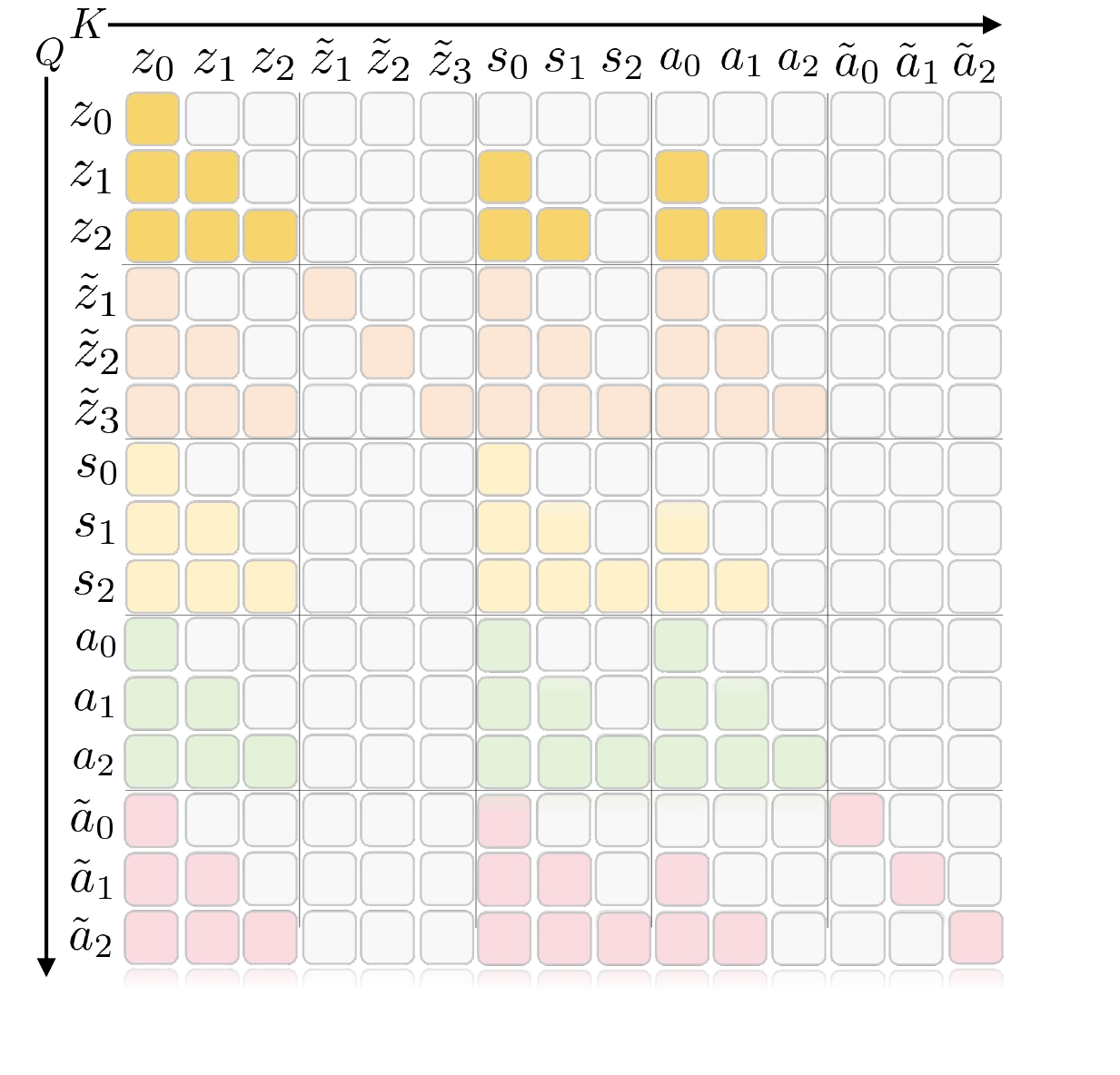}
    \vspace{-36pt}
    \caption{Teacher Forcing Attention Mask.}
    \label{fig:forcing}
\end{wrapfigure}

\paragraph{Embodiment-specific state and action interfaces.}
Robots differ in action dimensions, state dimensions, coordinate conventions, and semantic layouts.
Riemann-1.0 therefore shares the transformer backbone while specializing the input and output interfaces for each embodiment.
For an embodiment \(e\), raw action and state vectors are first converted to a canonical order defined by that embodiment.
They are then padded to fixed model dimensions \(d_a\leq D_a\) and \(d_s\leq D_s\), where \(D_a\) and \(D_s\) denote the maximum action and state dimensions.
The raw embodiment ID is mapped to a compact category slot, which indexes category-specific linear projections for action and state tokens.
The action prediction head is category-specific as well.
As a result, the model shares temporal and visual reasoning across embodiments while preserving the numerical action and state spaces of each robot embodiment.

\paragraph{Token sequence and temporal alignment.}
For each training example, Riemann-1.0 serializes all modalities into a single transformer sequence:
[\text{context latent}, \text{clean latent}, \text{noisy latent}, \text{state}, \text{clean action}, \text{noisy action}].
The context latent corresponds to the first observed frame.
Clean latent and clean action tokens serve as teacher-forced history under the causal mask, while noisy latent and noisy action tokens are the flow-matching targets to be denoised.
The action stream is temporally denser than the visual latent stream.

\paragraph{Fully causal training objective.}
Each visual latent summarizes a short segment of low-level control steps, determined by the VAE temporal stride and the frame sampling rate. 
During training, the action sequence is grouped by latent so that every predicted visual latent is paired with the corresponding chunk of robot actions. 
The state sequence contains one more element than the latent sequence, namely the initial state and the state observed after each latent transition. 
This organization aligns the training supervision with the causal rollout performed during inference.

To match online execution, each prediction should not access information unavailable at inference time.
Riemann-1.0 therefore employs a structured attention mask instead of unrestricted self-attention.
Clean tokens attend to previously observed clean tokens under the causal mask.
Within a local generation block, bidirectional attention can be used to model intra-block structure.
Noisy target tokens attend to preceding clean tokens and to other tokens within the same noisy block, but not to future clean tokens.
This prevents leakage from future observations while preserving teacher forcing over the observed trajectory prefix.

The attention mask is constructed from three token attributes: a frame ID, a time ID, and a clean/noisy flag. As shown in Figure~\ref{fig:forcing}, clean tokens attend only to clean tokens from the current or preceding time steps. Noisy tokens attend to clean tokens from preceding time steps and to noisy tokens within the same frame. Padding tokens, including text padding tokens, are masked out to ensure stable training with variable-length sequences.

\paragraph{Latent and action flow-matching losses.}
Riemann-1.0 applies the flow-matching objective to two prediction heads.
The visual-latent head predicts the velocity of noisy latent tokens:
\begin{equation}
\mathcal{L}_{z}
=
\mathbb{E}_{z,\epsilon,\sigma}
\left[
\left\|
v_\theta^z
\!\left(
z_\sigma,\sigma,
z_{<t},s_{<t},a_{\leq t},c
\right)
-
(\epsilon_z-z)
\right\|_2^2
\right].
\end{equation}
The action head predicts the velocity of noisy action tokens:
\begin{equation}
\mathcal{L}_{a}
=
\mathbb{E}_{a,\epsilon,\sigma}
\left[
\left\|
v_\theta^a
\!\left(
a_\sigma,\sigma,
z_{<t},s_{<t},a_{<t},c
\right)
-
(\epsilon_a-a)
\right\|_2^2
\right].
\end{equation}
The final objective balances visual dynamics and action prediction:
\begin{equation}
\mathcal{L}
=
(1-\lambda)\mathcal{L}_{z}
+ \lambda \mathcal{L}_{a},
\end{equation}
where \(\lambda\) controls the relative weight of the action objective.

\paragraph{Loss masking.}
Separate validity masks are applied to the latent and action losses. The latent mask excludes padded frames and invalid latents at episode boundaries. The action mask excludes padded action channels, padded frames, and invalid low-level action steps, with the loss normalized only over valid channel-step entries rather than the entire padded tensor. This masking strategy is essential for multi-embodiment training, where different robots occupy different subsets of the shared padded action space.

The same validity information is applied during noise injection.
Inactive padded regions are kept at zero for the clean sample, injected noise, noisy sample, and velocity target.
This avoids a train-test mismatch in which padded channels would contain random Gaussian noise during training but zeros during deployment.

\begin{wrapfigure}{l}{0.65\textwidth}
    \centering
    \vspace{-10pt}
    \includegraphics[width=0.65\textwidth]{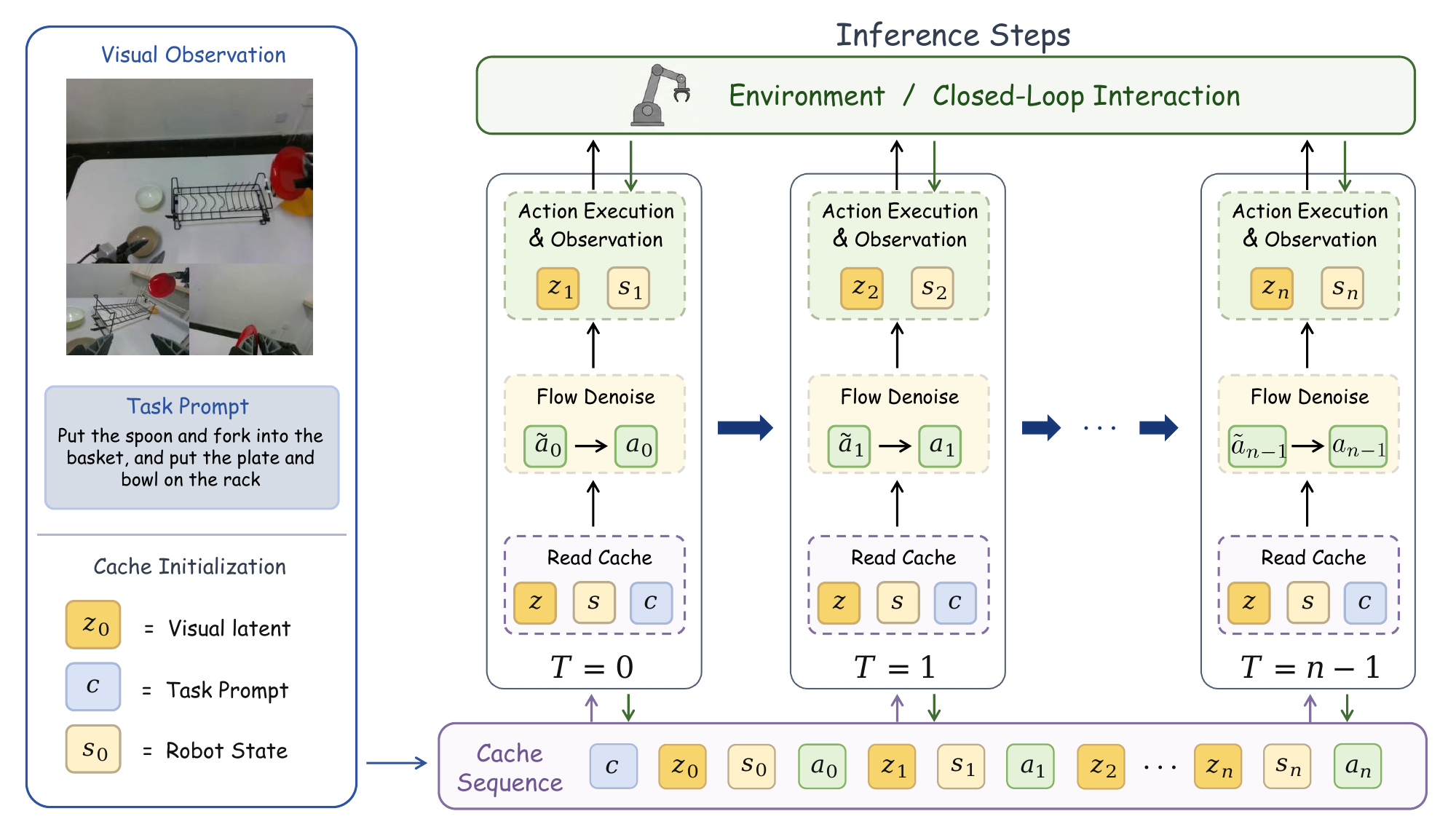}
    \caption{Autoregressive online inference of Riemann-1.0.}
    \vspace{-10pt}
    \label{fig:inference}
\end{wrapfigure}

\paragraph{Autoregressive online inference.}
The training attention mask naturally enables online autoregressive inference with a growing KV cache.
At the start of an episode, the first packed multi-view frame is encoded into \(z_0\), and the initial state \(s_0\) is embedded as a clean state token.
Together with the text embedding, these tokens initialize the cache. The autoregressive online inference process is illustrated in Figure~\ref{fig:inference}.

At denoising step \(t\), the model samples a Gaussian noisy action tensor and denoises it using the flow scheduler while attending to the cached clean history and text condition.
The resulting clean action chunk \(a_t\) is appended to the cache and executed for \(\mathrm{apf}\) low-level control steps.
The environment then returns new multi-view observations corresponding to the executed actions and the final state.
The observations then is packed, encoded into \(z_t\), and appended together with \(s_t\) to the cache as clean context for the next step.

This procedure follows Eq.~\ref{eq:unified_ar_wam}, where action generation precedes the corresponding visual latent update.
During real-robot execution, the next latent is obtained by encoding the actual observation returned by the environment.
During simulation, the same causal context together with the generated action chunk is fed to the latent head, and the predicted visual latent is decoded as the simulated camera observation.
For long-horizon rollouts, the cache is maintained within a fixed window.
Once the window is full, the cache is reset and the most recent observation becomes the new context frame, enabling continuous online control while preserving causal consistency.

\section{Training Recipe}

\begin{figure}
    \centering
    \includegraphics[width=0.75\textwidth]{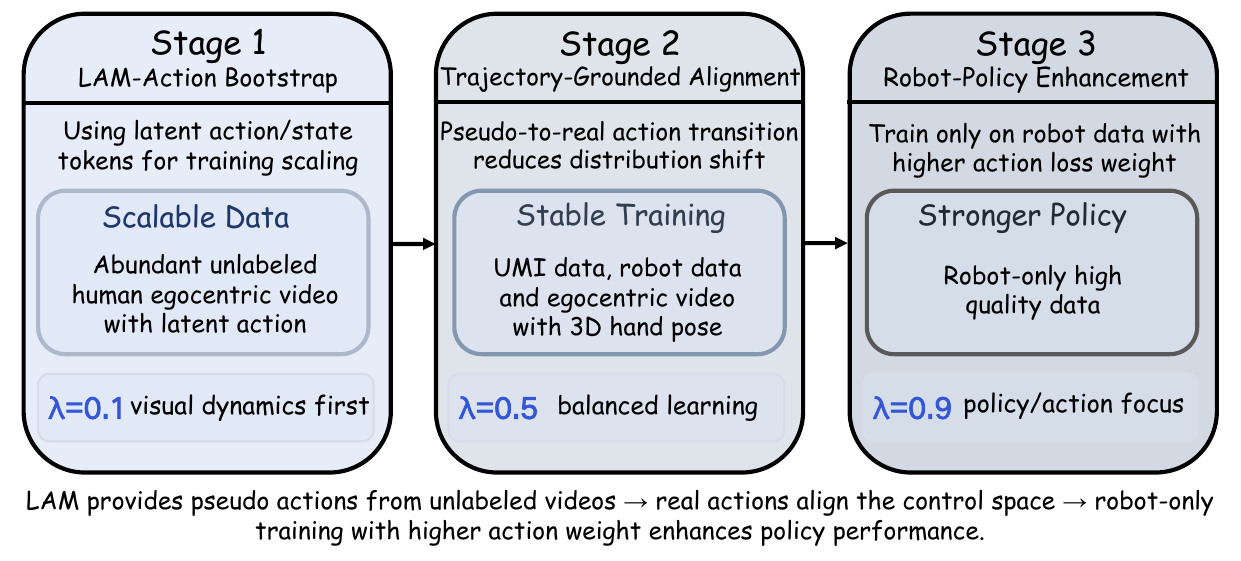}
    \caption{Three-stage WAM pretraining curriculum, from LAM-Action Bootstrap to Trajectory-Grounded Alignment and Robot-Policy Enhancement.}
    \label{fig:train_recipe}
\end{figure}

Rima-WAM is trained with a three-stage curriculum that progressively shifts supervision from video-only pseudo actions to real action trajectories.
This design is motivated by a practical data imbalance: video-only data is much larger and visually diverse, but it lacks executable robot actions, whereas real action datasets are smaller but provide direct supervision over robot behavior.
The first stage uses a frozen Latent Action Model (LAM) to train large-scale unlabeled videos with pseudo-action supervision.
The second stage grounds the model on a mixed corpus of UMI demonstrations, robot trajectories, and human videos with 3D hand annotations.
The third stage continues on high-quality robot-only data to sharpen executable control.
Across all stages, the model optimizes the same action-video objective,
\[
\mathcal{L}=(1-\lambda)\mathcal{L}_{z}+\lambda\mathcal{L}_{a},
\]
where \(\mathcal{L}_{z}\) is the visual-latent velocity loss and \(\mathcal{L}_{a}\) is the action velocity loss.
The action weight is increased from \(\lambda=0.1\) to \(0.5\) and then to \(0.9\), so the recipe first teaches broad visual dynamics with LAM-derived latent actions, then aligns the model with real multi-embodiment trajectories, and finally specializes it for robot execution. The overall three-stage pretraining framework is illustrated in Figure~\ref{fig:train_recipe}.

\subsection{Pretraining}

\paragraph{Stage I: LAM-Action Bootstrap.}
The first stage uses only unlabeled human manipulation videos.
Since these videos provide rich visual interaction but lack executable action labels, we first train a Latent Action Model (LAM)~\citep{routray2025vipra} and then freeze it as a pseudo-action annotator.
This stage aims to expose WAM to a large amount of manipulation dynamics before real robot action labels becomes available.
LAM provides this bridge by forcing the motion between two frames to pass through a compact latent bottleneck, so that the inferred latent code captures the underlying visual transition that an action would have induced and can serve as pseudo action supervision for WAM training.

LAM is trained as a latent-action VAE over pairs of adjacent video frames.
Given two frames \((x_t,x_{t+1})\), the encoder patchifies each frames into \(16\times16\) image patches and prepends a learnable action-prompt token to the token sequence of each frame.
A spatio-temporal transformer processes the two-frame token sequence, allowing the prompt token at the future frame to attend to both the current visual state and the resulting visual transition.
The hidden state of this future-frame action-prompt token is projected to Gaussian moments \((\mu,\log\sigma^2)\), defining a posterior distribution over a 32-dimensional latent action space.
A latent action is sampled with the reparameterization trick and injected into a decoder, which receives the current-frame visual patches and reconstructs the next frame.
This architecture prevents the decoder from observing \(x_{t+1}\) directly; to minimize reconstruction error, the latent bottleneck must encode the transition information needed to explain how \(x_t\) evolves into \(x_{t+1}\).
The LAM objective combines next-frame reconstruction and a weak KL regularizer:
\[
\mathcal{L}_{\mathrm{LAM}}
=
\|\hat{x}_{t+1}-x_{t+1}\|_2^2
+\beta D_{\mathrm{KL}}\!\left(q_\phi(z_t\mid x_t,x_{t+1})\,\|\,\mathcal{N}(0,I)\right).
\]
The KL term is deliberately weak: its role is to regularize the latent distribution without collapsing the transition code into an overly generic prior.
After training, we use the posterior mean \(\mu\) rather than a sampled latent as the deterministic pseudo action, which eliminating sampling noise when annotating large-scale video corpora.

During WAM pretraining, the frozen LAM is applied densely over each video window.
For every pair of adjacent frames in the dense stream, LAM predicts one pseudo transition code.
These codes are then grouped into action chunks that align with the WAM temporal interface: each visual latent corresponds to a short chunk of low-level action-like transitions.
In parallel, the same clip is temporally subsampled and encoded by the Wan VAE into visual latents, with the first latent serving as context and the remaining latents serving as future visual targets.
Thus Stage I presents WAM with the same structured inputs it will see later with real robot data, namely visual latents, state-like tokens, and dense action-like chunks, even though the original videos contain no action annotations.
This stage is not intended to learn a deployable robot policy; with \(\lambda=0.1\), it mainly initializes the visual dynamics backbone and teaches the model how action-like motion tokens correlate with future visual change.

\paragraph{Stage II: trajectory-grounded alignment.}
The second stage replaces pseudo actions with action trajectories from a mixed multi-embodiment corpus.
The data includes UMI demonstrations, robot trajectories, and human videos annotated with 3D hand poses and keypoints.
These sources provide more reliable action supervision absent from video-only data while still covering diverse viewpoints, objects, tasks, and embodiments.
Each trajectory is converted to the same WAM temporal interface: visual frames are packed according to the embodiment's camera layout, resized to the model resolution, and encoded into VAE latents; actions are grouped into dense chunks aligned with each visual latent; and states consist of the initial state plus one state after every latent transition.
For heterogeneous robots, action and state vectors are mapped to an embodiment-specific canonical order, normalized with per-embodiment statistics, and padded into the shared tensor format used by the model.
This stage uses \(\lambda=0.5\), making action prediction a central training signal while retaining visual-latent prediction as a dynamics regularizer.
Its goal is to ground the autoregressive action-video interface in real trajectory supervision and align the shared backbone across different camera layouts, action spaces, and robot bodies.
Compared with Stage I, the supervision is no longer only a latent explanation of visual motion; it now includes trajectories that can be interpreted as end-effector motion, gripper commands, joint commands, or hand-pose changes depending on the embodiment.
This makes the action stream increasingly policy-relevant while preserving the broad visual dynamics learned from video-only data. 

\paragraph{Stage III: robot policy enhancement.}
The final pretraining stage uses only high-quality robot demonstrations with real action-state supervision.
Human video data, LAM pseudo actions, UMI trajectories, and 3D hand-pose human trajectories are excluded.
This stage sharpens the model's executable action distribution after the previous stages have established broad visual dynamics and cross-embodiment trajectory alignment.
We increase the action weight to \(\lambda=0.9\), so optimization focuses primarily on state-conditioned action prediction and trajectory consistency, while the latent objective continues to regularize the learned dynamics.
The resulting model preserves the action-conditioned visual prediction ability of WAM, but its policy action head is more strongly biased toward robot-executable behavior.
In practice, this stage reduces the mismatch between broad pretraining data and downstream robot deployment: the model has already learned how actions and observations interact, and the robot-only stage calibrates this prior toward real actuator conventions, contact-rich execution, and policy stability.

\subsection{Post-training}

After pretraining, we post-train Riemann-1.0 to adapt the general manipulation prior to downstream tasks.
The post-training setting includes both real-world deployment tasks and simulation benchmarks.
For real-world evaluation, we collect teleoperation data on four tasks.
In \textit{ordered cube stacking}, the robot is required to stack four colored cubes in a specified color order, testing precise manipulation and instruction following.
In \textit{clothes folding}, the robot must manipulate a deformable object, testing whether the pretrained dynamics prior can adapt to non-rigid interaction.
In \textit{desk organization}, the robot tidies a table containing tissues, toys, chargers, tape, blocks, and pens, which requires long-horizon sequencing and fine-grained object handling.
In \textit{kitchen organization}, the robot places spoons, forks, bowls, and plates onto a kitchen rack, testing long-horizon manipulation in a cluttered household scene.
For each real-world task, we collect three hours of teleoperated demonstrations.
Instead of training task-specialized individual models, we aggregate all collected demonstration data to jointly fine-tune a single generalist model capable of executing all tasks.
In addition, compared with the pretraining stage, we increase the action loss weight to 0.95 in post-training, which improves action execution performance.

\section{Experiments}

\subsection{Real-World Experiments}

\subsubsection{Post-Training Data and Setup}

After large-scale pretraining, we further adapt Riemann-1.0 to real-world deployment through real-robot post-training. We build four representative manipulation scenarios on the Tianji Marvin dual-arm robot: ordered cube stacking, clothes folding, desk organization, and kitchen organization. These tasks cover order-constrained rigid-object stacking, deformable-object manipulation, long-horizon tabletop rearrangement, and long-horizon kitchen storage, forming a compact but challenging suite for household robot deployment. We collect 15 demonstrations for each task using human teleoperation.

In \textbf{ordered cube stacking}, the robot follows a language instruction to identify colored cubes and assemble a four-layer stack following a specified color order.  Four-layer stacking sharply increases the difficulty: small alignment errors at lower layers are amplified by later placements, and a single misaligned layer can destabilize the entire stack. Therefore, this task requires accurate visual recognition of object color and pose, planning of the grasping sequence, and precise pick-and-place execution.

In \textbf{clothes folding}, the robot manipulates a deformable garment rather than a rigid object. Clothes can wrinkle, stretch, and drift in pose during contact, so the robot must continuously adapt its action strategy based on the observed state. This task stresses deformable-object perception, bimanual coordination, and fine-grained contact control.

In \textbf{desk organization}, the robot must identify and rearrange multiple scattered desktop objects into their designated locations. A key challenge is inserting slender objects such as pens into a pen holder. This requires accurate perception of the pen pose and the holder opening, exploiting the rim of the holder to reorient the pen toward a vertical pose, and then controlling the gripper to complete a precise insertion.

In \textbf{kitchen organization}, the robot must complete an end-to-end  tidying task in a kitchen scene. It first localizes and grasps utensils such as forks and spoons, then places them back into the target storage area. It must also grasp yellow and red plates and insert them vertically into a drying rack, stack bowls, and place the bowl stack on the upper layer of the rack, requiring long-horizon sequencing and stable placement under cluttered conditions.

We compare Riemann-1.0 with representative open-source VLA and WAM baselines, including DreamZero, $\tau_0$-WM, LingBot-VLA, $\pi_{0.5}$, LingBot-VA, and G0.5, under the same real-world manipulation setting. For DreamZero, we reproduce the model using the official open-source implementation, training a 5B-scale model on our collected large-scale manipulation dataset, denoted as DreamZero*.

\subsubsection{Real-World Evaluation}

We evaluate real-robot performance using two metrics. Success Rate (SR) measures whether the final task goal is completed. Progress Success Rate (PSR) measures the fraction of task progress completed according to task-specific intermediate milestones, which is especially important for long-horizon organization tasks where partial completion is meaningful even when the final state is not fully reached.

\begin{table*}[t]
    \centering
    \caption{Real-world manipulation results. SR and PSR are reported in percent.  DreamZero* denotes our reproduced implementation based on the official open-source code, trained with our large-scale manipulation dataset.}
    \label{tab:realworld_results}
    \resizebox{\linewidth}{!}{%
    \begin{tabular}{l*{5}{cc}}
        \toprule
        \multirow{2}{*}{Model} & \multicolumn{2}{c}{Ordered cube stacking} & \multicolumn{2}{c}{Kitchen organization} & \multicolumn{2}{c}{Clothes folding} & \multicolumn{2}{c}{Desk organization} & \multicolumn{2}{c}{Avg.} \\
        \cmidrule(lr){2-3} \cmidrule(lr){4-5} \cmidrule(lr){6-7} \cmidrule(lr){8-9} \cmidrule(lr){10-11}
        & SR & PSR & SR & PSR & SR & PSR & SR & PSR & SR & PSR \\
        \midrule
        DreamZero*~\citep{ye2026dreamzero} & 15.0 & 20.0 & 15.0 & 16.6 & 15.0 & 12.5 & 15.0 & 33.3 & 15.00 & 20.60 \\
        \(\tau_0\)-WM~\citep{zhou2026tau0wm} & 15.0 & 26.2 & 15.0 & 20.0 & 15.0 & 24.6 & 20.0 & 57.2 & 16.25 & 32.00 \\
        LingBot-VLA~\citep{wu2026lingbotvla} & 15.0 & 31.5 & 20.0 & 80.2 & 80.0 & 88.4 & 20.0 & 76.9 & 33.75 & 69.25 \\
        \(\pi_{0.5}\)~\citep{intelligence2025pi_} & 40.0 & 59.0 & 20.0 & 37.2 & 45.0 & 73.5 & 40.0 & 73.1 & 36.25 & 60.70 \\
        LingBot-VA~\citep{li2026lingbotva} & 40.0 & 66.6 & 20.0 & 30.0 & 80.0 & 85.0 & 40.0 & 80.1 & 45.00 & 65.43 \\
        G0.5~\citep{galaxea2026g05} & 80.0 & 89.5 & 35.0 & 47.8 & \textbf{85.0} & 90.0 & \textbf{80.0} & 93.4 & 70.00 & 80.18 \\
        \midrule
        Riemann-1.0 (Ours) & \textbf{85.0} & \textbf{91.6} & \textbf{90.0} & \textbf{98.4} & \textbf{85.0} & \textbf{92.5} & \textbf{80.0} & \textbf{95.2} & \textbf{85.00} & \textbf{94.43} \\
        \bottomrule
    \end{tabular}
    }
\end{table*}

As shown in Table~\ref{tab:realworld_results}, Riemann-1.0 achieves the best average performance among all compared models, with an average SR of 85.00\% and an average PSR of 94.43\%. It maintains at least 80.0\% SR on all four tasks and exceeds 91.0\% PSR on every task, indicating that the model not only reaches the final goal more reliably but also makes stable progress through intermediate steps. These results suggest that post-training preserves the general manipulation prior learned during pretraining while adapting it to the target real-world robot, camera layout, and household task distribution.

\begin{figure}[htbp]
    \centering
    \vspace{-4mm}
    \includegraphics[width=0.95\linewidth]{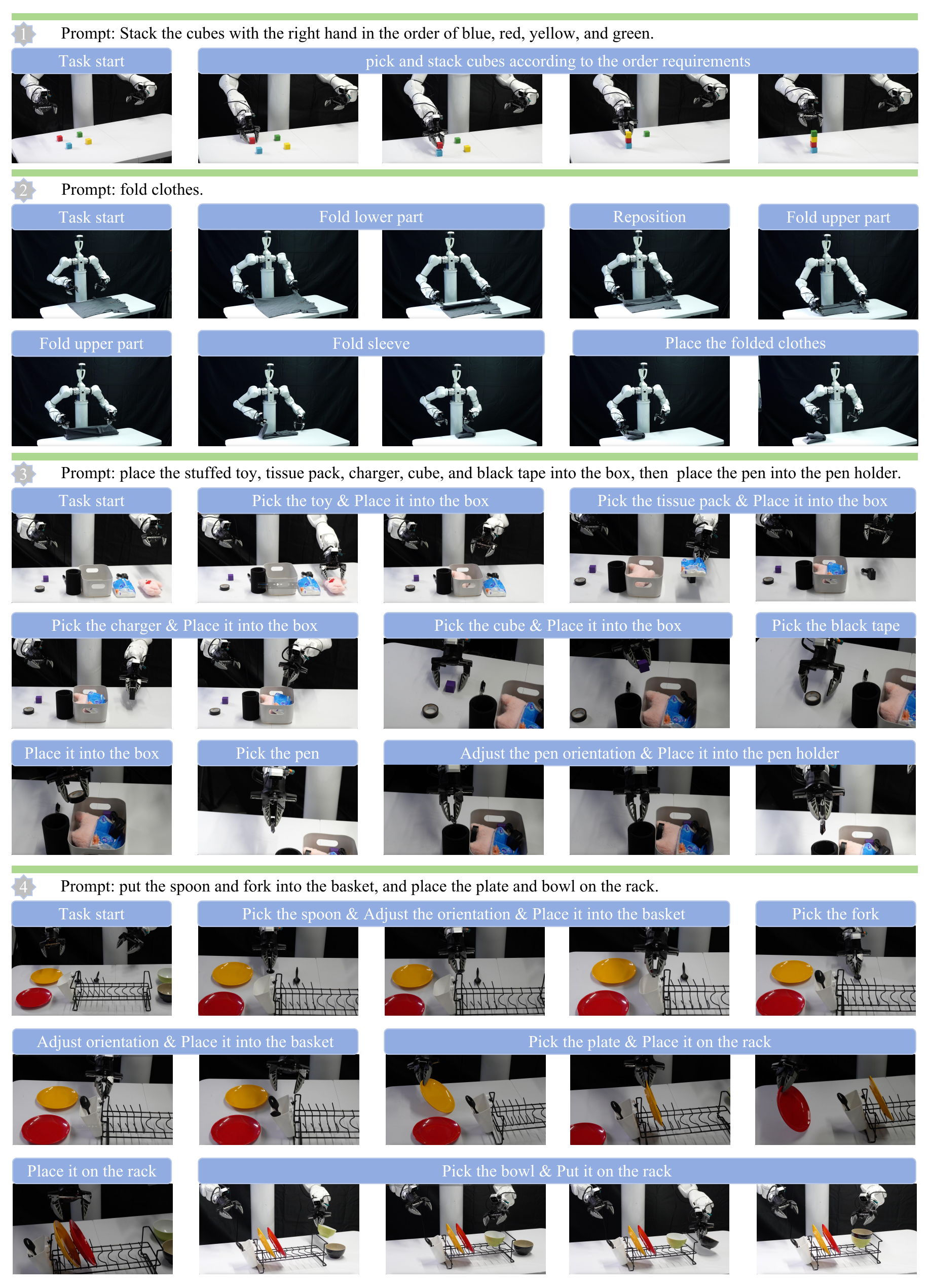}
    \vspace{-4mm}
    \caption{Robot execution processes for real-world tasks; clothes folding shows a single folding sequence.}
    \label{fig:real}
\end{figure}

Specifically, on ordered cube stacking, Riemann-1.0 reaches 85.0\% SR and 91.6\% PSR, improving over the strongest baseline G0.5 by 5.0 SR points and 2.1 PSR points. On kitchen organization, Riemann-1.0 obtains 90.0\% SR, while the next best SR is 35.0\% from G0.5; it also improves PSR over the strongest baseline LingBot-VLA by 18.2 points, indicating more reliable completion of long-horizon tableware storage. On clothes folding, Riemann-1.0 ties the best SR at 85.0\% with G0.5 but achieves the highest PSR at 92.5\%, suggesting better intermediate progress on deformable-object manipulation. On desk organization, Riemann-1.0 ties the best SR at 80.0\% with G0.5 and further improves PSR from 93.4\% to 95.2\%, reflecting more stable execution when inserting slender objects and arranging cluttered desktop items. The execution process of the robot on these tasks is illustrated in Figure~\ref{fig:real}.

\begin{figure}[htbp]
    \centering

    \includegraphics[width=0.95\linewidth]{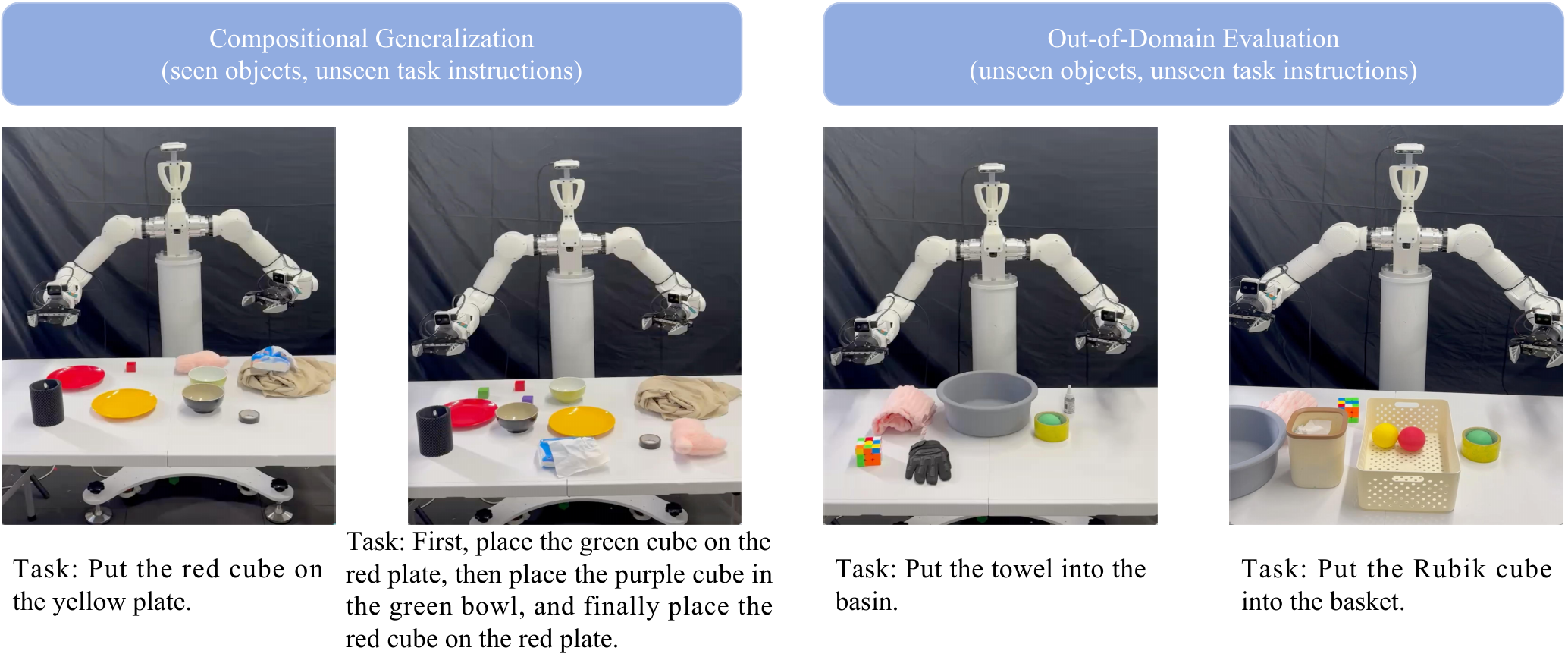}
  
    \caption{Illustration of compositional generalization and OOD tasks.}
    \label{fig:ood}
\end{figure}

\subsubsection{Compositional Generalization and Out-of-Domain Evaluation}

Beyond evaluating the four tasks used for real-world post-training, we further test whether the deployed policy can reuse its learned manipulation skills under held-out instructions. As shown in Figure \ref{fig:ood}, we consider two setting. 

\textbf{Compositional generalization} uses seen object categories but recombines them into unseen task goals. In the first task, the robot is instructed to grasp a specified cube and place it into a target receptacle, such as a bowl or a plate. This setting keeps the object category familiar while changing the object-receptacle relation. In the second task, the instruction further specifies object color, receptacle color, receptacle type, and execution order, e.g., \textit{First, place the green cube on the red plate, then place the purple cube in the green bowl, and finally place the red cube on the red plate}. 

\textbf{Out-of-Domain (OOD) evaluation} uses unseen scenes, objects, and task goals that are not included in the real-world post-training set. No additional demonstrations or parameter updates are provided for these tasks. The task definitions are: placing a Rubik's cube into a storage box and placing a towel into a basin. Since each task has a binary task completion criterion, we report Success Rate (SR) over 10 real-robot trials per task.

\begin{table*}[t]
    \centering
    \small
    \caption{Real-world compositional generalization and OOD results. Task rows report successful trials out of 10.}
    \label{tab:realworld_generalization_ood}
    \begin{tabular}{llccc}
        \toprule
        Setting & Task & Riemann-1.0 (Ours) & \(\pi_{0.5}\) & LingBot-VA \\
        \midrule
        \multirow{3}{*}{Compositional} 
        & Cube-to-bowl/plate placement & \textbf{8/10 (80.0)} & 6/10 (60.0) & 3/10 (30.0) \\
        & + color-order constraint & \textbf{5/10 (50.0)} & 3/10 (30.0) & 0/10 (0.0) \\
        \cmidrule(lr){2-5}
        & Average & \textbf{65.0} & 45.0 & 15.0 \\
        \midrule
        \multirow{3}{*}{OOD}
        & Rubik's cube to storage box & \textbf{10/10 (100.0)} & 5/10 (50.0) & 3/10 (30.0) \\
        & Towel-to-basin placement & \textbf{7/10 (70.0)} & 5/10 (50.0) & 3/10 (30.0) \\
        \cmidrule(lr){2-5}
        & Average & \textbf{85.0} & 50.0 & 30.0 \\
        \midrule
        Overall average & & \textbf{75.0} & 47.5 & 22.5 \\
        \bottomrule
    \end{tabular}
\end{table*}

As shown in Table~\ref{tab:realworld_generalization_ood}, Riemann-1.0 achieves the highest SR in both held-out settings. In compositional generalization, it succeeds in 8 out of 10 trials when placing cubes into a bowl or plate. When the instruction additionally specifies object color, receptacle color, receptacle type, and execution order, the task becomes harder because the policy must jointly solve visual grounding and multi-step sequencing. Riemann-1.0 still reaches 50.0\% SR under this stricter setting, leading to an average SR of 65.0\% across the two compositional tasks.

In the OOD setting, Riemann-1.0 obtains an average SR of 85.0\%. The model completes all Rubik's-cube-to-storage-box trials and retains 70.0\% SR on towel-to-basin placement, showing that the policy can execute both rigid-object storage and deformable-object placement under held-out scenes and instructions. Across all four held-out tasks, Riemann-1.0 obtains an overall average SR of 75.0\%. These results suggest that the pretrained manipulation prior and real-world post-training do not merely fit the training tasks, but transfer to new task compositions and out-of-distribution household scenarios.

\subsection{Simulation Evaluation}
\label{sec:sim_bench}

We evaluate Riemann-1.0 on three simulation benchmarks: RoboCasa365, RoboTwin 2.0, and LIBERO. 
Following the standard protocol of each benchmark, we report task success rate as the primary metric. 
We keep the same WAM temporal interface across benchmarks: one visual latent corresponds to 16 low-level action steps.

\subsubsection{RoboCasa365}

For RoboCasa-365~\citep{nasiriany2026robocasa365}, we first pretrain on 300 tasks and then finetune on the Target-50 split. 
The 50 target tasks evaluation contains Atomic-Seen, Composite-Seen, and Composite-Unseen groups. 

\begin{table}[htbp]
\centering
\caption{Evaluation results on RoboCasa365 50 target tasks.}
\label{tab:benchmark_robocasa365}
\begin{tabular}{lcccc}
\toprule
\textbf{Model} & \textbf{Atomic-Seen} & \textbf{Composite-Seen} & \textbf{Composite-Unseen} & \textbf{Average} \\
\midrule
GR00T-N1.5~\citep{bjorck2025grootn1} & 60.6 & 35.0 & 33.3 & 43.7 \\
Fast-WAM~\citep{yuan2026fastwam} & 59.1 & 36.4 & 33.2 & 43.5 \\
LingBot-VA~\citep{li2026lingbotva} & 63.5 & 37.3 & 32.1 & 45.1 \\
ABot-M0.5~\citep{chen2026abot} & 70.6 & 44.3 & 45.6 & 54.2 \\
\midrule
Riemann-1.0 (Ours) & \textbf{74.2} & \textbf{56.0} & \textbf{56.3} & \textbf{62.6} \\
\bottomrule
\end{tabular}
\end{table}

As shown in Table~\ref{tab:benchmark_robocasa365}, Riemann-1.0 achieves the best performance across all three RoboCasa365 categories, improving the average success rate from 54.2\% to 62.6\% over the strongest baseline, ABot-M0.5. Notably, Riemann-1.0 improves the performance on Composite-Seen and Composite-Unseen by 11.7 and 10.7 percentage points, respectively, demonstrating its strong compositional generalization capability and ability to transfer learned manipulation skills to unseen tasks.

\begin{table}[htbp]
\centering
\caption{Evaluation results on the RoboTwin 2.0 benchmark.}
\label{tab:benchmark_robotwin}
\begin{tabular}{lccc}
\toprule
\textbf{Model} & \textbf{Clean} & \textbf{Randomized} & \textbf{Average} \\
\midrule
$\pi_{0.5}$~\citep{intelligence2025pi_} & 82.7 & 76.8 & 79.8 \\
ABot-M0~\citep{yang2026abot} & 86.1 & 85.1 & 85.6 \\
LingBot-VLA~\citep{wu2026lingbotvla} & 86.5 & 85.3 & 85.9 \\
Qwen-VLA~\citep{wang2026qwen}& 86.1 & 87.2 & 86.7 \\
Motus~\citep{bi2026motus} & 88.7 & 87.0 & 87.8 \\
Fast-WAM~\citep{yuan2026fastwam} & 91.9 & 91.8 & 91.8 \\
LingBot-VA~\citep{li2026lingbotva}& 92.9 & 91.6 & 92.2 \\
G0.5~\citep{galaxea2026g05} & 93.7 & 92.8 & 93.3 \\
ABot-M0.5~\citep{chen2026abot}&94.0& \textbf{94.2} &94.1 \\
\midrule
Riemann-1.0 (Ours) & \textbf{94.6} & 94.0 & \textbf{94.3} \\
\bottomrule
\end{tabular}
\end{table}

\subsubsection{RoboTwin 2.0}

For RoboTwin 2.0~\citep{chen2025robotwin}, we train on 50 bimanual tasks. 
The training set contains clean scenes with 50 demonstrations per task, plus 25,000 demonstrations from heavily randomized scenes, i.e., 500 randomized demonstrations per task. 
Evaluation is reported under Clean and Randomized settings to measure both nominal performance and robustness to visual and physical perturbations.  As reported in Table~\ref{tab:benchmark_robotwin}, Riemann-1.0 achieves the highest average success rate on RoboTwin 2.0, reaching average 94.3\% across 50 bimanual tasks.

\begin{table}[H]
\centering
\caption{Evaluation results on the LIBERO benchmark.}
\label{tab:benchmark_libero}
\begin{tabular}{lccccc}
\toprule
\textbf{Method} & \textbf{Spatial} & \textbf{Object} & \textbf{Goal} & \textbf{Long} & \textbf{Average} \\
\midrule
$\pi_0$-Fast~\citep{pertsch2025fast} & 96.4 & 96.8 & 88.6 & 60.2 & 85.5 \\
GR00T-N1~\citep{bjorck2025grootn1} & 94.4 & 97.6 & 93.0 & 90.6 & 93.9 \\
$\pi_0$~\citep{black2024pi0} & 98.0 & 96.8 & 94.4 & 88.4 & 94.4 \\
InternVLA-M1~\citep{chen2025internvla} & 98.0 & 99.0 & 93.8 & 92.6 & 95.9 \\
$\pi_{0.5}$~\citep{intelligence2025pi_} & 98.8 & 98.2 & 98.0 & 92.4 & 96.9 \\
GR00T-N1.6~\citep{bjorck2025grootn1} & 97.7 & 98.5 & 97.5 & 94.4 & 97.0 \\
OpenVLA-OFT~\citep{kim2025fine}& 97.6 & 98.4 & 97.9 & 94.5 & 97.1 \\
Fast-WAM~\citep{yuan2026fastwam} & 98.2 & \textbf{100.0} & 97.0 & 95.2 & 97.6 \\
Motus~\citep{bi2026motus} & 96.8 & 99.8 & 96.6 & 97.6 & 97.7 \\
LingBot-VA~\citep{li2026lingbotva} & 98.5 & 99.6 & 97.2 & 98.5 & 98.5 \\
ABot-M0~\citep{yang2026abot} & 98.8 & 99.8 & 99.0 & 96.6 & 98.6 \\
Being-H0.5~\citep{luo2026being} & 99.2 & 99.6 & \textbf{99.4} & 97.4 & 98.9 \\
G0.5~\citep{galaxea2026g05} & 98.4 & \textbf{100.0} & 98.6 & \textbf{98.6} & 98.9 \\
\midrule
Riemann-1.0 (Ours) & \textbf{99.6} & \textbf{100.0} & 97.6 & \textbf{98.6} & \textbf{99.0} \\
\bottomrule
\end{tabular}
\end{table}

\subsubsection{LIBERO}

For LIBERO~\citep{liu2023libero}, we train on four official suites: Spatial, Object, Goal, and Long. Each suite contains 10 tasks, and each task provides 50 demonstrations.

On LIBERO, as shown in Table~\ref{tab:benchmark_libero}, Riemann-1.0 reaches the  overall average of 99.0\%.  Specifically, it achieves 99.6\% on Spatial, 100.0\% on Object, 97.6\% on Goal, and 98.6\% on Long.

\begin{figure}[H]
    \centering
    \includegraphics[width=\linewidth]{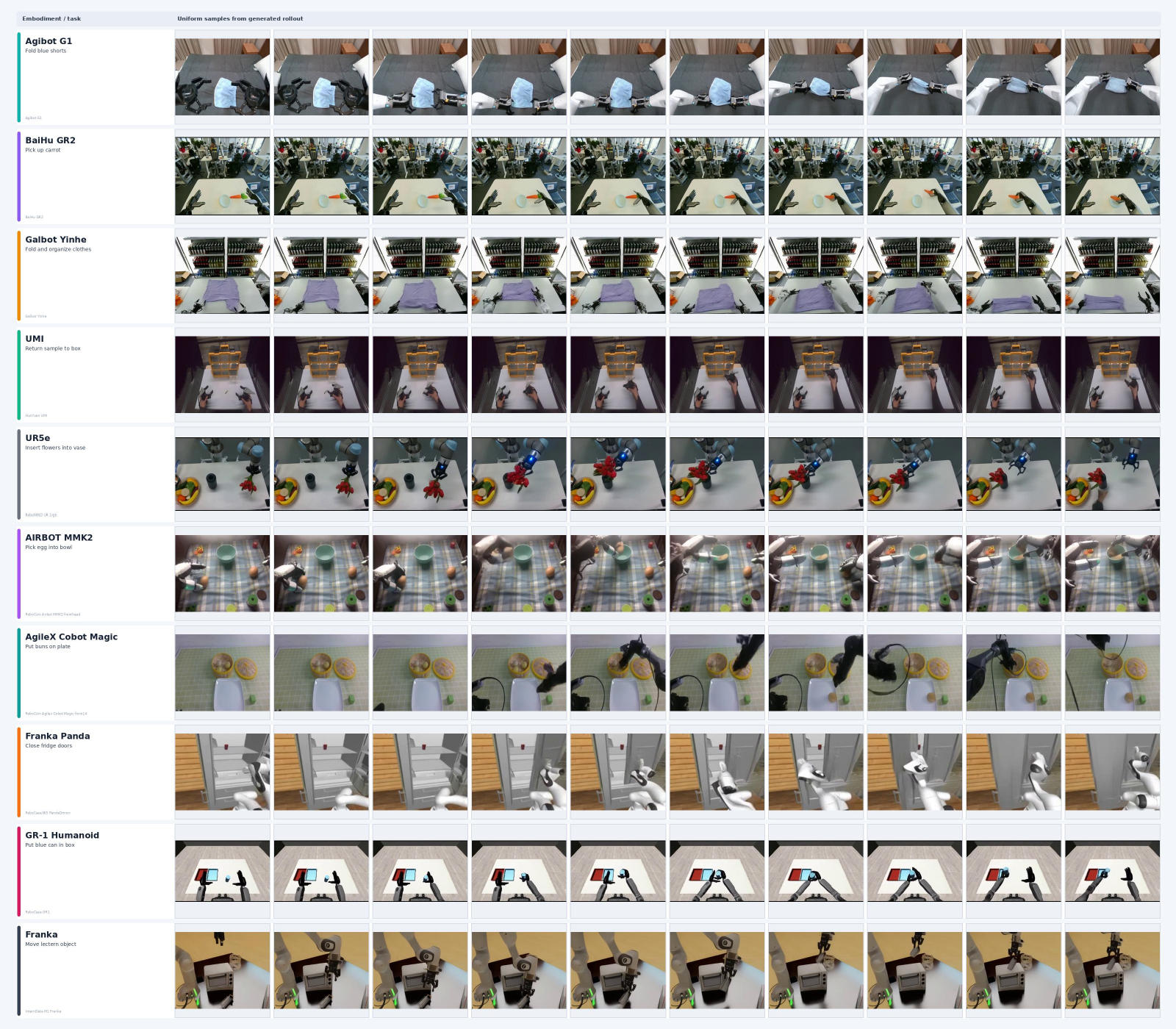}
    \caption{\textbf{Multi-embodiment action-conditioned visual rollout.}
    Generated multi-embodiment rollouts from Riemann-1.0. Each row presents a rollout for a different robot embodiment with an abbreviated task description, demonstrating its capability as a data-driven simulator across diverse robots, viewpoints, and manipulation tasks.
    }
    \label{fig:multi_embodiment_visual_rollout}
\end{figure}

\subsection{Action-Conditioned Visual Rollout as a Multi-Embodiment Simulator}

Riemann-1.0 can also be used as an action-conditioned visual simulator. In this mode, the model is not asked to only output an executable action. Instead, it receives the current visual observation, task prompt, robot state, and a candidate future action trajectory, and predicts the visual consequences of executing the trajectory. The action trajectory is represented with the same embodiment-specific action interface used during policy training, then embedded as action tokens and inserted into the causal action-video sequence. These action tokens condition the visual-latent denoising process, so the generated frames are tied to the robot motion rather than sampled as an unconditional future video.

At inference time, we first encode the current observation into the VAE latent space and use it as the clean visual context. The future action chunk is either provided by the policy head, sampled from a candidate plan, or taken from a recorded trajectory for analysis. Conditioned on the prompt, state, clean visual context, and action tokens, the latent head rolls out future visual latents; these latents are then decoded back into RGB video. For longer horizons, the generated or observed visual output can be appended back to the context and the same procedure can be repeated autoregressively. This gives Riemann-1.0 a data-driven simulator interface: actions are treated as controllable inputs, while future camera observations are produced as the simulated consequences.

\Cref{fig:multi_embodiment_visual_rollout} shows qualitative rollouts across heterogeneous embodiments and camera layouts. The examples cover humanoid robots, dual-arm systems, single-arm platforms, dexterous-hand robots, and simulation benchmarks such as RoboTwin~\citep{mu2025robotwin} and RoboCasa-GR1~\citep{bjorck2025grootn1}. Despite the differences in viewpoint, embodiment geometry, gripper appearance, and task type, the generated videos preserve the main scene structure and exhibit action-consistent object and robot motion. These results suggest that Riemann-1.0 learns a shared action-conditioned visual dynamics prior that can generalize beyond a single robot body, making it useful for rollout inspection, action-plan comparison, and future-state imagination.

\section{Related Work}

\paragraph{Vision-language-action policies.}
Large vision-language-action (VLA) policies have become a central paradigm for general robot manipulation.
RT-1 and RT-2 demonstrate that transformer-based policies can benefit from broad robot demonstrations and web-scale semantic pretraining~\citep{brohan2022rt1,brohan2023rt2}, while OpenVLA, \(\pi_0\), GR00T-N1, and LingBot-VLA further scale open backbones, continuous action prediction, cross-embodiment data, and system-level robot learning pipelines~\citep{kim2024openvla,black2024pi0,bjorck2025grootn1,wu2026lingbotvla}.
These models are effective executable policies, but they are usually optimized as observation-to-action predictors rather than explicit predictors of the visual consequences of actions.
At the same time, scaling embodied learning requires heterogeneous data sources.
Egocentric human video datasets provide large-scale hand-object interactions and long-horizon activity structure~\citep{damen2020rescaling,grauman2023egoexo4d,wang2023holoassist,liu2022hoi4d,liu2024taco,hoque2025egodex}, UMI and exoskeleton-glove demonstrations provide manipulation-centric views and device states closer to robot control~\citep{chi2024umi,tao2025dexwild}, and robot trajectory datasets provide direct action-state supervision but are fragmented across embodiments, camera layouts, and control conventions~\citep{openx2023,robomind2024,agibotworld2025,interndataA1_2025,galaxea2025openworld,robocoin2026}.
Riemann-1.0 is designed to use these sources jointly: human videos contribute scalable interaction knowledge, UMI-style data bridges human manipulation and robot-compatible control, and robot trajectories provide embodiment-specific executable supervision.

\paragraph{World action models.}
World models learn predictive dynamics for planning, evaluation, or policy learning; in robot manipulation, this direction has recently evolved into World Action Models (WAMs), which jointly model robot actions and future observations.
Existing WAMs differ mainly in how they order action and video generation.
DreamZero-style methods couple action and video through joint denoising~\citep{ye2026dreamzero}; LingBot-VA emphasizes native video-action pretraining on top of video generative models~\citep{li2026lingbotva}; FastWAM studies the efficiency trade-off of using video modeling during training while keeping inference action-centered~\citep{yuan2026fastwam}; and GIGA-World-Policy distinguishes future-visual supervision during training from efficient action decoding during inference~\citep{ye2026gigaworld}.
Related controllable world-modeling systems also study action-conditioned video rollout and future-aware policy evaluation~\citep{guo2025ctrlworld,zhou2026tau0wm}.
Beyond architecture, recent robot learning systems increasingly rely on staged training recipes, combining semantic pretraining, robot imitation learning, action-space alignment, data filtering, balancing, normalization, and downstream fine-tuning~\citep{kim2024openvla,black2024pi0,bjorck2025grootn1,wu2026lingbotvla,openx2023,robomind2024,agibotworld2025,tread2026}.
Riemann-1.0 follows this trend but adapts it to WAM pretraining: it first learns from unlabeled human videos using LAM-derived pseudo actions, then grounds the action-video interface with mixed UMI, robot, and 3D hand-pose trajectories, and finally strengthens executable behavior with robot-only policy enhancement.

\paragraph{Policies and action-conditioned simulators.}
Most deployed robot policies are optimized for fast action prediction, while model-based simulators are optimized for predicting future observations or state transitions.
Classical model-based control separates these roles, but video-based robot learning has increasingly blurred the boundary by using video prediction for planning, future representation learning, and policy supervision~\citep{du2023unipi,du2023vlp,bruce2024genie,cheang2024gr2,hu2024vpp,liang2025videopolicy,shen2025videovla}.
The key requirement for a useful robot simulator is action conditioning: generated future observations should respond to the robot's proposed motion rather than extrapolating a likely video without control input.
Riemann-1.0 addresses this requirement with a fully causal autoregressive formulation.
Actions are predicted or provided before the corresponding visual latent, the action conditions the visual transition, and the resulting observation becomes context for the next step.
Thus, when deployed as a policy, the model predicts the next action chunk from causal history and uses the real environment observation as the next context; when used as a simulator, the same action chunk conditions the latent head to generate the corresponding future visual observation.
This dual interface connects executable control and model-based imagination within a single WAM.

\section{Conclusion}
We introduced Riemann-1.0, a fully causal autoregressive World Action Model for embodied intelligence. Riemann-1.0 addresses two central challenges in scaling robot foundation models: how to unify heterogeneous embodied experience across egocentric human videos, handheld-gripper demonstrations, and robot trajectories, and how to jointly model robot actions, states, and world evolution in a causal form aligned with real interaction. By combining a unified embodied data infrastructure with Progressive Embodied Pretraining over more than \num{200}K+ hours of embodied experience, Riemann-1.0 transfers large-scale interaction knowledge from weak supervision to executable robot control.
Across simulation and real-world evaluations, Riemann-1.0 achieves state-of-the-art performance, including 62.6\% on RoboCasa365, 94.3\% on RoboTwin 2.0, 99.0\% on LIBERO, and 85.0\% SR with 94.43\% PSR on long-horizon real-world manipulation tasks. These results show that Riemann-1.0 provides a scalable path for transforming large-scale embodied experience into generalizable robot manipulation capabilities.

\clearpage
\bibliography{main}
\clearpage

\end{document}